\documentclass[conference]{IEEEtran}
\IEEEoverridecommandlockouts
\usepackage{cite}
\usepackage{amsmath,amssymb,amsfonts}
\usepackage{algorithmic}
\usepackage{graphicx}
\usepackage{textcomp}
\usepackage{xcolor}
\usepackage{multicol}
\usepackage{caption}
\usepackage{xfrac}
\usepackage{booktabs} % Latex tables from pandas
\usepackage[utf8]{inputenc}
\usepackage{caption,subcaption}
\def\BibTeX{{\rm B\kern-.05em{\sc i\kern-.025em b}\kern-.08em
    T\kern-.1667em\lower.7ex\hbox{E}\kern-.125emX}}
\begin{document}

\title{Vessel and Port Efficiency Metrics through Validated AIS data}

\author{\IEEEauthorblockN{Tomaž Martinčič\IEEEauthorrefmark{1}\IEEEauthorrefmark{2},
Dejan Štepec\IEEEauthorrefmark{1}\IEEEauthorrefmark{2}, Joao Pita Costa\IEEEauthorrefmark{1}, Kristijan Čagran\IEEEauthorrefmark{1} and Athanasios Chaldeakis\IEEEauthorrefmark{3}}
\IEEEauthorblockA{\IEEEauthorrefmark{1}XLAB Research,
Ljubljana, Slovenia\\}
\{tomaz.martincic, dejan.stepec, joao.pitacosta, kristijan.cagran\}@xlab.si
\IEEEauthorblockA{\IEEEauthorrefmark{2}University of Ljubljana,
Faculty of Computer and Information Science, Ljubljana, Slovenia\\}
\IEEEauthorblockA{\IEEEauthorrefmark{3}Piraeus Port Authority,
Piraeus, Greece\\}
}

\maketitle

\begin{abstract}
Automatic Identification System (AIS) data represents a rich source of information about maritime traffic and offers a great potential for data analytics and predictive modelling solutions, which can help optimizing logistic chains and reducing environmental impacts. In this work, we address the main limitations of the validity of AIS navigational data fields, by proposing a machine learning-based data-driven methodology to detect and (to the possible extent) also correct erroneous data. Additionally, we propose a metric that can be used by vessel operators and ports to express numerically their business and environmental efficiency through time and spatial dimensions, enabled with the obtained validated AIS data. We also demonstrate Port Area Vessel Movements (PARES) tool, which demonstrates the proposed solutions.
\end{abstract}

\begin{IEEEkeywords}
AIS, data validation, port metrics, environment, covid-19
\end{IEEEkeywords}

\section{Introduction}

Automatic Identification System (AIS) data represents a rich source of  maritime traffic information, and offers a great potential for data analytic and predictive modelling solutions, which can help optimizing logistic chains and reducing environmental impacts. In this work we address the main limitations of the validity of AIS navigational data fields, by proposing a machine learning-based data-driven methodology to detect and (to the most possible extent) also correct erroneous data. Additionally, we propose a metric that can be used by vessel operators and ports to numerically express their business and environmental efficiency through time and spatial dimensions, enabled with such obtained validated AIS data.

\begin{figure*}[htbp]
    \centering
    \begin{subfigure}{0.49\textwidth}
      \includegraphics[width=\linewidth]{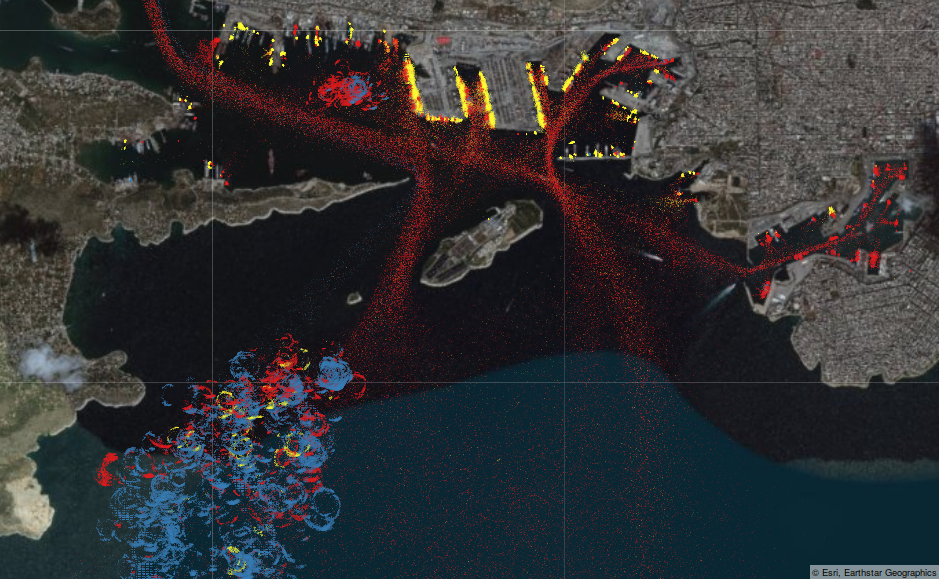}
      \caption{}
      \label{fig:AIS_NAVSTAT_1b}
    \end{subfigure}\hfil
        \begin{subfigure}{0.49\textwidth}
      \includegraphics[width=\linewidth]{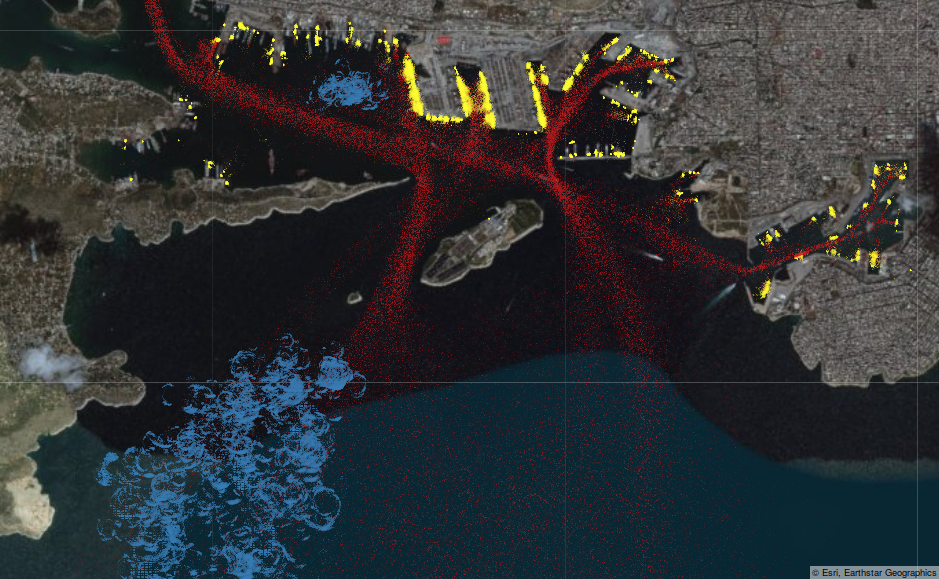}
      \caption{}
      \label{fig:PIXEL_NAVSTAT_1}
    \end{subfigure}\hfil
    \caption{Reported AIS navigational statuses with errors (a) and corrected navigational statues by our methods (b). Red: under way using engine, blue: at anchor, yellow: moored.}
    \label{fig:ppa_navstat}
\end{figure*}

Different studies have reported significant errors in the transmitted AIS data~\cite{AIS_error}, which can significantly reduce its usability and effectiveness for gathering insights using automated data analytics solutions. One of the most interesting AIS reported fields is that of navigational status, which could serve as an essential tool to measure port's business (e.g. vessel waiting times, vessel turnaround times), as well as environmental (e.g. emission) metrics. Studies have found that at least 30\% of the vessels were detected as transmitting incorrect status information. An example of errors in AIS navigational status is presented in Figure~\ref{fig:AIS_NAVSTAT_1b}. Vessels can be moored only at terminals. However, sometimes they report moored navigational status in the anchorage areas, or even while sailing. Other navigational statuses are also occasionally falsely reported.

Some of the static AIS reported fields (e.g. vessel type, MMSI, IMO), can be corrected through dedicated maritime vessel databases, but this is not the case with the dynamically reported fields (e.g. navigational status). In order to correct navigational status in real-time, we propose and compare three different approaches. The first approach utilizes the vessel's speed and location, along with the predetermined location of anchorage and terminal areas. The downside of this method is the requirement of setting the speed threshold and the availability of GIS data (anchorage and port areas) from the ports. As the second approach, we propose to utilize the vessel's speed and rotation information, which omits the need for manually provided data. Finally, we propose to utilize different machine learning techniques~\cite{clustering1,catboost} on reported AIS fields (spatial and kinematic), which automatically classifies reported information into appropriate navigational status, that is validated with the reported one. An example of corrected AIS transmitted navigational data is presented in Figure~\ref{fig:PIXEL_NAVSTAT_1}.

Validated AIS data is used to extract vessel voyages in a predefined area, and the time spent in different navigational positions is calculated, in order to evaluate port entry and port turnaround times. We have evaluated our proposed methods in the Port of Piraeus (Greece), where historical data (one year) and live AIS data is collected from the AISHub\footnote{http://www.aishub.net/}. Vessel arrivals and turnaround times are validated against FAL forms data collected by the port. Proposed port efficiency metrics based on waiting times are globally applicable, without the need for manually provided data. Port efficiency metrics for different ports are computed in an interval of one year and combined with vessel-specific AIS reported fields (e.g. vessel type, vessel length), which provides additional value to the proposed automatic data-driven metric. This allows for the comparison on a vessel and cargo level.

Our main contributions can be summarized as follows:

\begin{itemize}  
    \item Three different methods for AIS navigational status validation and correction, based on static and manually provided data, as well as a completely automatic machine learning-based approach.
    
    \item The procedure for extracting vessel voyages out of validated AIS data in a predefined area, together with port and vessel efficiency metric for the calculation of time spent in different navigational positions (e.g. port entry waiting time, port turnaround time).
    
    \item The evaluation of the proposed methods on live and historical AIS data in different ports, together with validation against ground truth data provided in Port of Piraeus (FAL forms).
\end{itemize}

\section{Related Work}

The availability of large-scale Automatic Identification System (AIS) data, made available by the recent build-up of terrestrial, as well as satellite constellations offers an essential tool for maritime situational awareness applications~\cite{ais_analysis}. Different studies have reported significant errors in the transmitted AIS data~\cite{AIS_error}, which can significantly reduce its usability and effectiveness for gathering insights using automated data analytics solutions. One of the most interesting AIS reported fields is that of navigational status, which could serve as a tool to measure port's business (e.g. vessel waiting times, vessel turnaround times), as well as environmental (e.g. emission) metrics. Studies have found that at least 30\% of the vessels were detected as transmitting incorrect status information. Most of the work is focused on using reported positional data for route extraction, prediction and anomaly detection~\cite{ais_analysis, ais_anomaly, ais_prediction} and are not utilizing the reported AIS navigational statuses, due to its unreliability~\cite{ais_emissions}.

Most of the work around port areas and regions of high-density traffic is focused on simple statistical analysis and visualizations of reported AIS data~\cite{ais_istanbul, ais_singapore}. In our work, we propose to use validated AIS data to extract vessel voyages in a predefined area and the time spent in different navigational positions, in order to evaluate port entry and port turnaround times. Besides serving as a useful tool to measure aspects of the port's business metric (e.g. vessel waiting times, vessel turnaround times), the same approach also supports the estimation of environmental (e.g. emission) impacts~\cite{ais_emissions}. In comparison with the most common approach of AIS navigational data use and validation~\cite{ais_emissions, ais_emissions2}, where speed information is used to classify navigational status into three newly defined classes (cruising mode, maneuvering mode, hotelling mode), we propose a method that validates, preserves and uses the originally reported navigational statuses.

\section{Port Area Vessel Movements - PARES}

\begin{figure*}[ht!]
    \centering
    \includegraphics[width=\linewidth]{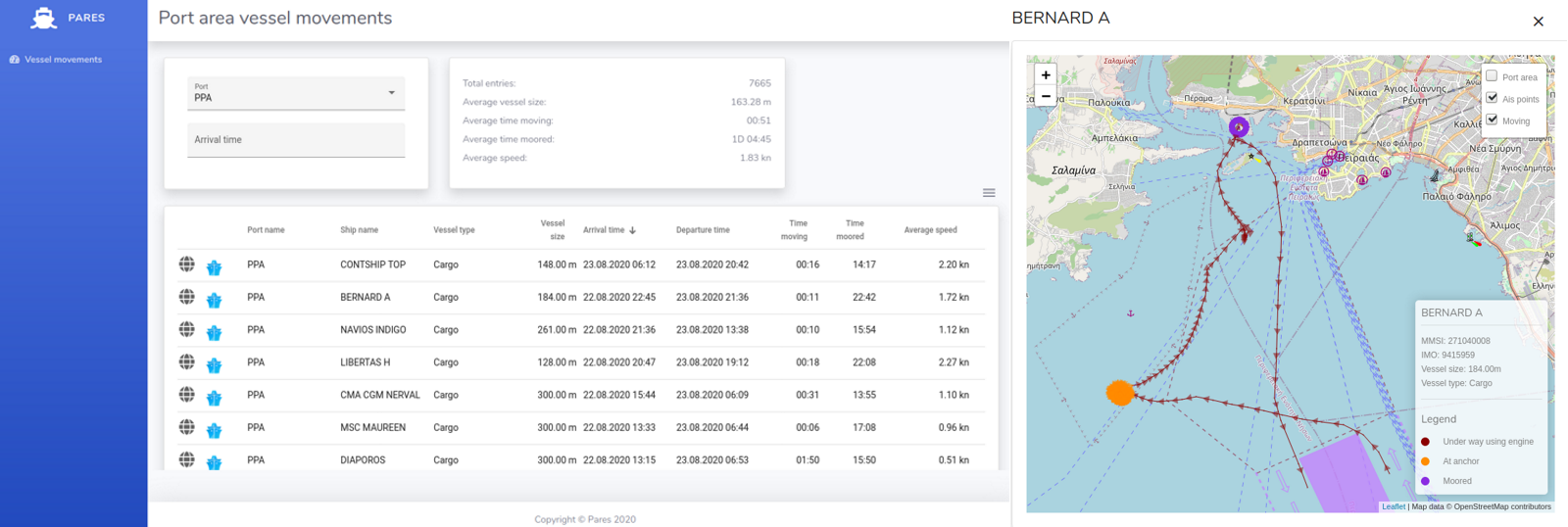}
    \caption{Port Area Vessel Movements (PARES) tool used for maritime traffic analysis around Port of Piraeus using validated AIS data, depicting vessels captured in the bay area and detailed analysis for a particular vessel.}\label{fig:pares_banner}
\end{figure*}

The Port Area Vessel Movements (PARES) tool (presented in Figure~\ref{fig:pares_banner}), validates and utilizes AIS data to extract information about vessels movements inside the port areas. The tool can process historical data or work in live mode. The process of extracting movements and times consists of three main steps. In the first step, the PARES tool has to select data based on the selected time period, the vessel's location and the vessel types. In the second step, the model has to validate the data and extract additional features, which are used in the third step to clean the data and group the AIS messages into separate voyages. In the context of PARES, voyage is defined as all movements that vessels make inside of the port area at single arrival to the port. The first vessel arrives to the port area, then it sails to the terminal with an optional stop in the anchorage area. When the vessel arrives to the terminal, the cargo processing operations can begin. After all cargo is unloaded and loaded onto the vessel, it leaves the port. The PARES model extracts information about waiting times at the anchorage area, cargo processing times at the terminal and the duration of the vessel movements inside of the port area, together with the average speed of the movements.

\subsection{AIS Data Validation}

% NAVSTAT
One of the most important features in AIS messages, also validated by the PARES is the vessel navigational status. This is a crucial piece of information, used to split the voyage into separate parts. The tool takes advantage of three navigational statuses: 0 (underway using engine), 1 (at anchor), and 5 (moored). We developed three different approaches to determinate vessel navigational status. Each approach has its own advantages and disadvantages.

\begin{figure*}[ht!]
    \centering
    \begin{minipage}[b]{.45\textwidth}
        \centering
         \includegraphics[width=\linewidth]{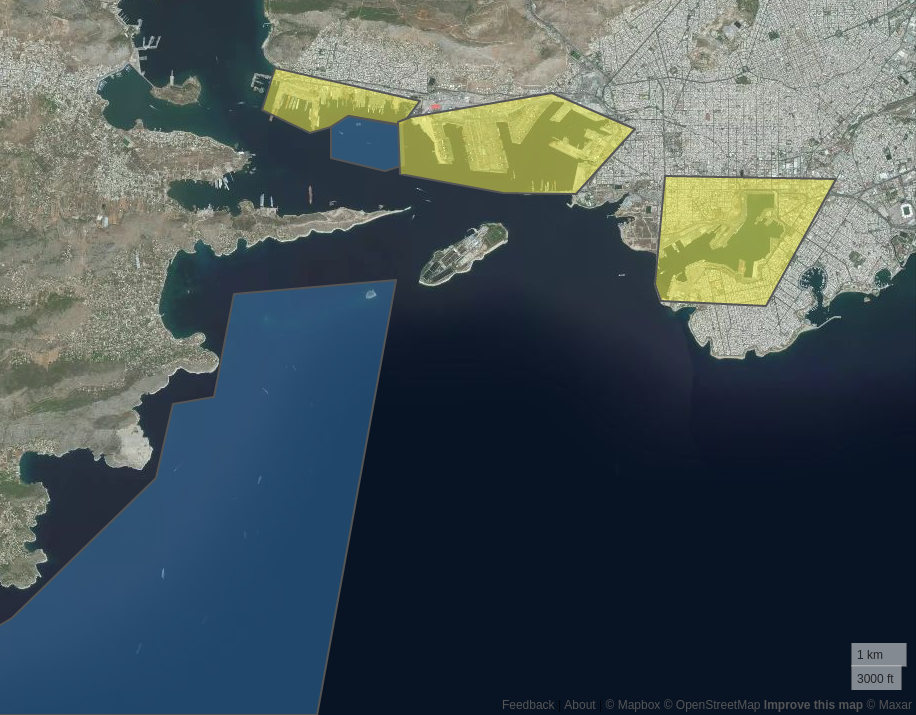}
        \caption{Manually drawn polygons in PPA port area. Anchorage area with blue color and terminals with yellow.}\label{fig:ais_ppa_polygons}
    \end{minipage}\qquad
    \begin{minipage}[b]{.45\textwidth}
        \centering
        \includegraphics[width=\linewidth]{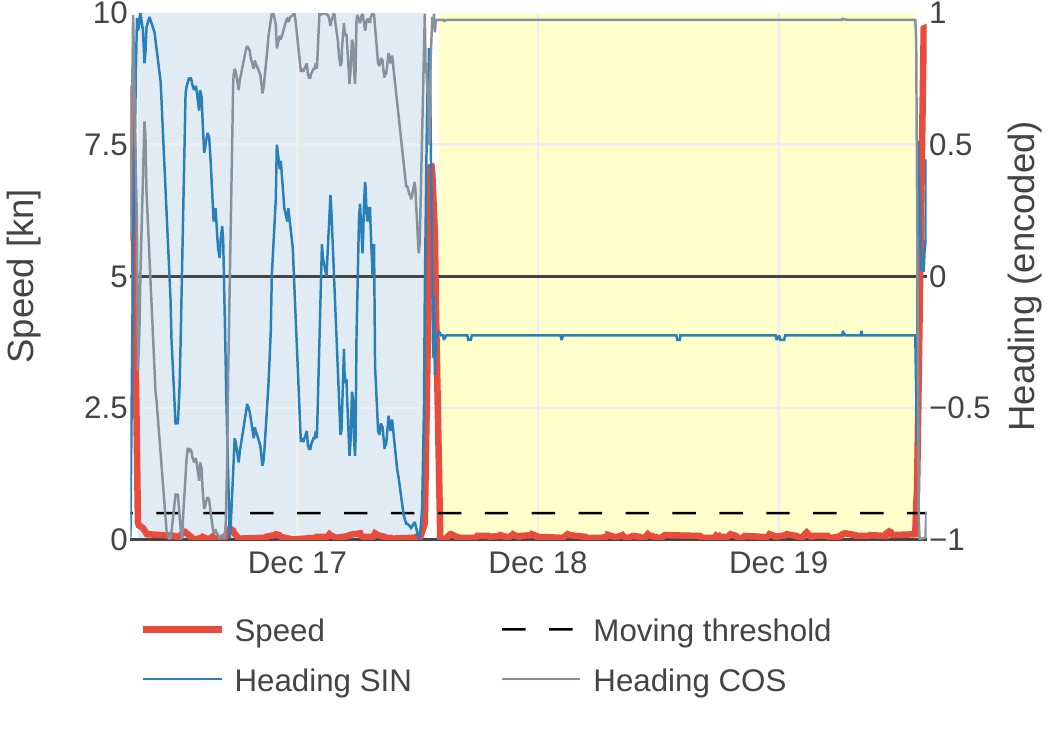}
        \caption{Vessel speed and its encoded heading during anchored (blue background) and moored (yellow background) period.}\label{fig:anchorage_and_moored_movements}
    \end{minipage}
\end{figure*}

%% NAVSTAT - Polygons
In the first approach navigational statuses are determined based on vessel location and its speed. For the location, we observe if the vessel is inside predefined polygons that mark anchorage and terminal areas (see Figure \ref{fig:ais_ppa_polygons}). The main problems here are that polygons have to be determined manually and that the vessel can anchorage on the border of a polygon. In that case, the PARES tool would falsely detect multiple changes of the navigational status. The vessel can be entering and exiting anchorage area polygon multiple times during single anchoring, due to normal vessel movements during anchorage.

%% NAVSTAT - Vessel movements
In the second approach we observe the vessel's speed and rate of rotation. Moored and anchored vessels are not moving, with exception of rotations of anchored vessels. This distinction can be used to separate moored and anchored vessels. For example, vessel's speed and heading, with marked anchoring and mooring periods, are plotted in Figure \ref{fig:anchorage_and_moored_movements}. Vessels that are moored at terminal cannot rotate and vessels at anchorage normally rotate due to winds and water flows (circles in anchorage area in the Figure \ref{fig:ppa_navstat}). The method is very accurate for longer stops, but can be unreliable for detecting short duration anchorages, where the vessel has not rotated enough. It is important to encode vessel heading, considering it has cyclical values in range between 0° and 360°. We encoded it using sin cos cyclical encoding ($\sin({\frac{2\pi heading}{360}})$ and $\cos({\frac{2\pi heading}{360}})$). This yields two features with values between -1 and 1.

%% NAVSTAT - ML
In the third approach we utilized machine learning algorithms. Multiple algorithms and libraries were evaluated. From unsupervised clustering technique HDBSCAN~\cite{clustering1}, to supervised classification methods, such as CatBoost~\cite{catboost} and k-nearest neighbours (KNN). With HDBSCAN we obtained clusters of anchored and moored vessels, from AIS messages of stopped vessels, based on their location and reported navigational status.  With CatBoost we tried to classify navigational status from vessel location and speed. We also prepared lagged features, that helped the model for more accurate classifications. The cleanest result was produced by KNN with large number of neighbours taken into account. With the KNN we classified navigational status of stopped vessels from vessel location. For best results, at least 300 neighbours were used.

% For training we used
% Messages are clustered based on vessel location, speed, rotation speed. For this method we need enough historical data. % tukaj opiši več

% Machine learning approach is used to identify, if vessel that is not moving is moored or anchored. To prepare training data set, we selected AIS messages with navigational status of 1 or 5, and speed lower than 0.5 kn. Only three features were used: longitude, latitude and reported navigational status in AIS message. 
% % Kater način predlagaš? Catboost ali clustering?

% In the third approach we utilized supervised gradient boosting machine learning algorithm packaged in CatBoost\footnote{\url{https://catboost.ai/}} python library. To automatically extract as clean as possible training data set, we filtered AIS data based on reported navigational status and speed. If vessel reported navigational status 0, its speed had to be more than X knots(TODO). If vessel reported navigational status 1 or 5, it should not be moving. Threshold for speed was set on Y knots (TODO). This filters removed AIS messages with obviously falsely reported navigational statuses. 
% The problem of this method is, that training data set has no data with speed in range between X and Y knots.

% Missing data
Another problem is that of the missing AIS messages. There are different types of data outages. If there are missing AIS messages of a single vessel: AIS transmitter is not working properly, or it is turned off on purpose. If there are missing AIS messages in a specific area, there is a problem with terrestrial based AIS receiver, or the area is not covered. If there is no AIS data for some period, the problem is in the AIS data provider or in our AIS data collection system. In some cases we can just ignore the data loss. If the vessel was moored or anchored for the whole time of data outage, or the vessel was moving in straight line. In other cases we cannot extract correct times. The best solution is to mark vessel stops or time periods with missing data. 

% single vessel --> vessel AIS transmitter
% single area --> shore based AIS receiver 
% whole system --> our AIS data collector
% or if we can interpolate vessel movement.

% There are multiple reasons for missing AIS data. For example: problems with AIS transmitter on vessel, problems with shore-based AIS receiver, AIS transmitter on vessel is on purpose turned off, problems with AIS stream collector.
% We are collecting AIS data from AISHub. There is no access to historical AIS data. In case of failure on our data collector, there is no option to get data for downtime period. 

\subsection{Voyage Extraction}
Voyage represents vessel movement in port area. From the arrival to the port area, the optional stop in the anchorage area, the stop at the terminal and the departure from the port area. % Duplicated

Validated messages are grouped into the voyages based on the vessel identification number (MMSI) and time between two consecutive AIS messages in the port area. If there is less than 24 hours between two messages from the same vessels in the port area, the messages can be grouped into the voyage. If the gap in messages is more than 5 hours long and the vessel moved for more than 100 meters in that time, the chain of AIS messages is split into two voyages at that point. 
% A to opišemo? Opišemo prav pandas? Mogoče idejo za streming?
To extract voyages, we built pandas dataframe from AIS messages in a selected port area. Each row represents a single AIS message, and columns represent AIS message fields. We sorted rows by MMSI and time. To split data in different voyages, we built a set of different conditions that compare two consecutive rows. Change of MMSI, more than 24 hour time gap, more than 5 hour time gap and change of location for more than 100 meters. For efficient comparison of consecutive rows, we utilized the pandas dataframe shift method.

% Extracting voyages from validated AIS data is pretty simple. The biggest problem is missing data. The set of rules is developed in order to handle missing data in different parts of voyages. 

% Voyages are assembled from up to 6 parts: arrival to port area, optional path to anchorage area and anchoring, path to terminal, mooring, departure from port area. Gaps in data can occur over any part of voyage and the handling of missing data has to be adapted to the part. 

% If the gap in the AIS data is longer than 24 hour, the next message is accounted as new voyage. 
% Conditionally gaps over 5 hours can also mark a new voyage, if vessel moved for more than 100 meters during the data outage.

\begin{figure*}[ht!]
    \centering
    \includegraphics[width=\linewidth]{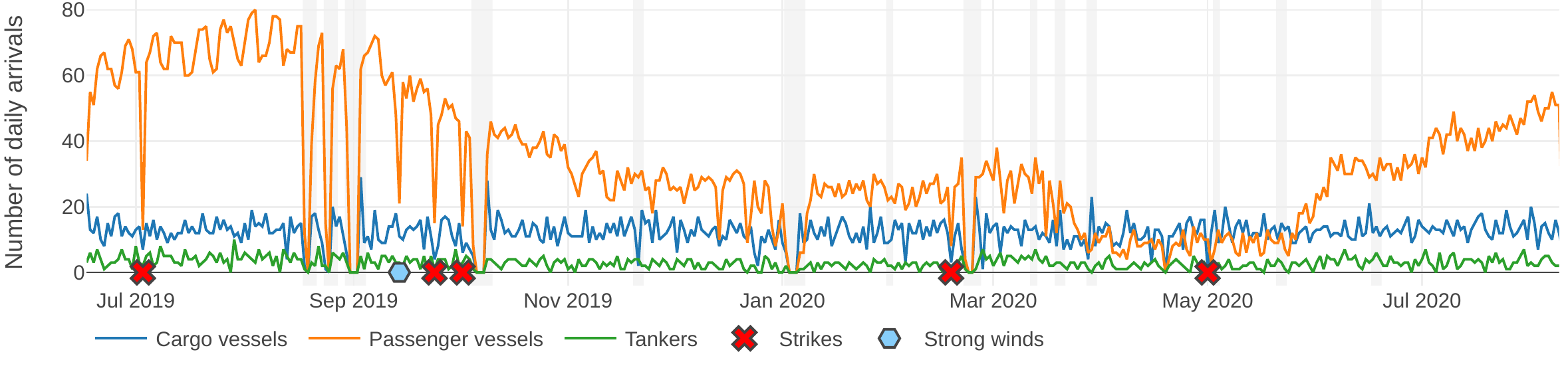}
    \caption{Number of daily arrivals of cargo vessels, passenger vessels and tankers to the Port of Piraeus. Different events are labelled on a timeline, representing the cause for the reduce traffic, Gray areas represents periods with missing AIS data.}\label{fig:ppa_daily_arrivals_cargo_tankers_passenger}
\end{figure*}

\subsection{Port Efficiency Metrics}

The PARES tool is capable of providing different metrics for describing the current state of vessels in the port and its surrounding areas, as well as historical data analysis.

Turnaround time is one of the ports' key performance indicators. The faster cargo is unloaded and loaded, the better the port is. The PARES can automatically extract this information from AIS data. Time periods aggregations (e.g. weekly aggregations) can be used to monitor the port's performance through time. It is also possible to compare different ports. We identified several vessels with multiple arrivals in two or more ports, for which we have enabled the PARES tool. One of the problems when comparing different ports is that we don't have information about amount of cargo.

It is possible to monitor arrival and departure times for different parts of port. With the tool someone could spot abnormal or even missing arrival or departure of regular vessels lines. 

% TODO is it possible for vessel

\section{Experiments}

The PARES tool was tested on more than one year of historical AIS data made available by AISHub and validated on 2 months of data collected in Port of Piraeus (Greece). Though, it is directly applicable to any port in the world, without the need of port cooperation. It is hard to get ground truth data for validation of the results from the PARES tool. For the validation of the turnaround times in the port, the AIS gathered data can be validated against the data by the Port Community Systems (PCS) in the ports. The data is manually entered, so we have no way of knowing how accurately the times are entered. It is possible to export vessel calls data from MarineTraffic\footnote{https://www.marinetraffic.com}, but for the ports of our interest, polygons capture larger port area and not just the terminals, thus turnaround times can have significant offsets of multiple hours, making data not accurate enough in order to validate the PARES tool.

\begin{figure}[ht!]
    \centering
    \includegraphics[width=\linewidth]{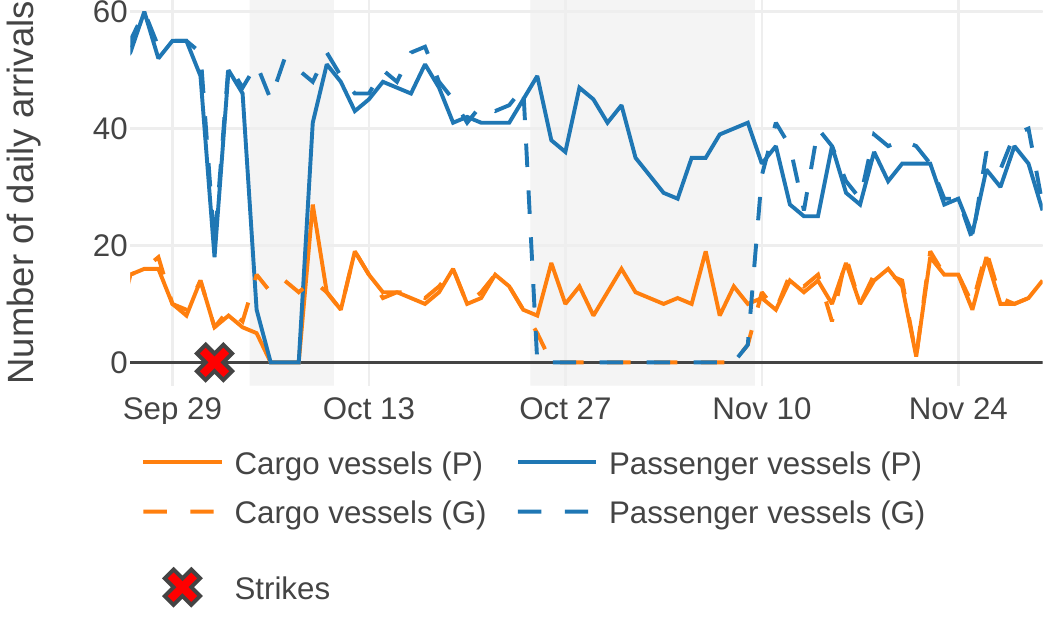}
    \caption{Number of daily arrivals detected by PARES compared with ground truth vessel call data from the port. Comparison is made separately for passenger and cargo vessels. Gray area depicts missing data in the port or in the AIS collection process.}\label{fig:ppa_daily_entrances_ground_truth_comaprison}
\end{figure}

Obtained vessel calls data from Port of Piraeus is missing vessel departure times and contains only arrival times. Therefore, we were not able to compare turnaround times detected by the PARES with the ground truth data captured in the port. Instead, we compared number of daily arrivals detected by the PARES with daily arrivals captured in the PCS. Averaged Mean Average Error (MAE) across all vessel types is 4.46 with an average of 54 arrivals per day, captured in the ground truth data. The comparison is plotted in Figure \ref{fig:ppa_daily_entrances_ground_truth_comaprison}.

Aggregated vessel arrivals provide useful information for analysis. Number of daily arrivals to the Port of Piraeus, based on vessel type, are presented in Figure \ref{fig:ppa_daily_arrivals_cargo_tankers_passenger}. We grouped vessel types into 3 different categories, cargo vessels (AIS type 70 - 79), tankers (AIS type 80 - 89) and passenger vessels (AIS type 40 - 49 and 60 - 69). The number of arrivals of cargo vessels and tankers has some variance, but no obvious trend. The passenger vessels follow a strong seasonal trend, with the most traffic happening in the summer months. In the summer of 2019, we can observe weekly periodical trends in passenger vessels traffic, with peaks over the weekends. Grey periods are marking the time with possible disruptions on the collection of AIS data. There are five labour strikes\footnote{https://citystarcar.com/strikes/}, most of them having visible noticeable effect on the number of daily arrivals, especially on passenger vessels. Extreme weather (e.g. strong winds) conditions can also interrupt the maritime traffic. The number of arrivals of passenger vessels severely dropped in March, due to the progress of the COVID-19 pandemic in Europe. The maritime passenger traffic started to recover in the end of May. However, the numbers in July and August are still lower than in the last year. No significant reduction of traffic can be noticed for cargo and tanker vessels in the Port of Piraeus.

\begin{table}
    % \scriptsize
    % \footnotesize
    \small
    \centering
    \caption{Arrivals and departures of High speed vessel (MMSI: ) into PPA passenger terminal.}
    \begin{tabular}{lll}
    \toprule
        Arrival time & Departure time & Turnaround time \\
    \midrule
        \multicolumn{1}{c}{$\cdots$} & \multicolumn{1}{c}{$\cdots$} & \multicolumn{1}{c}{$\cdots$} \\
        2019-09-12 14:28 &	2019-09-13 04:25 &	0 days 13:46 \\
        \textbf{2019-09-13 15:44} &	\textbf{2019-09-15 04:20} &	\textbf{1 days 12:26} \\
        2019-09-15 16:19 &	2019-09-16 04:18 &	0 days 11:49 \\
        2019-09-16 14:17 &	2019-09-17 04:22 &	0 days 13:49 \\
        \multicolumn{1}{c}{$\cdots$} & \multicolumn{1}{c}{$\cdots$} & \multicolumn{1}{c}{$\cdots$} \\
        2019-09-21 14:34 &	2019-09-22 04:19 &	0 days 13:34 \\
        2019-09-22 14:36 &	2019-09-23 04:22 &	0 days 13:35 \\
        \textbf{2019-09-23 13:59} &	\textbf{2019-09-25 04:21} &	\textbf{1 days 14:07} \\
        2019-09-25 14:11 &	2019-09-26 04:19 &	0 days 13:58 \\
        \multicolumn{1}{c}{$\cdots$} & \multicolumn{1}{c}{$\cdots$} & \multicolumn{1}{c}{$\cdots$} \\
    \bottomrule
    \end{tabular}
    \label{tbl:hs_vessel_arrivals_departures}
\end{table}

PARES also provides a possibility to explore the exact arrival departure times of a specific vessel. A subset of features, representing the results of the PARES tool for a single passenger vessel (HIGHSPEED 4, MMSI: 239658000) are presented in Table \ref{tbl:hs_vessel_arrivals_departures}. The vessel has a regular schedule, with daily departures at 4:20AM in the morning and arrivals around 2:10PM in the afternoon. It is possible to monitor possible delays in arrivals or departures. In the presented sample, the vessel did not depart from the passenger terminal in two occasions in a short period of time. The first missed departure was caused by strong winds\footnote{https://www.ferryhopper.com/en/blog/ferry-news/strong-winds-greek-ferries-latest} and the second one by a general ferry strike\footnote{https://news.gtp.gr/2019/09/17/ferry-strike-greece-announced-september-24/}.

\section{Conclusion}

In this work we present the Port Area Vessel Movements (PARES) tool which validates and extracts information about vessel movements inside the port areas. Several static and machine learning-based data-driven techniques were presented to detect and (to the possible extent) also correct erroneous data. Validated Automatic Identification System (AIS) data is used for data analytics and predictive modelling solutions, to which the considered use-cases present significant opportunities to optimize logistic chains and reduce environmental impacts. The proposed metrics can be used by vessel operators and ports to express numerically their business and environmental efficiency through time and spatial dimensions, when enabled with the obtained validated AIS data. These enhanced information can translate into improvements to the business intelligence at the port, allowing for better decision-making based on more accurate data insight at a low price in what regards data acquisition. 

\section*{Acknowledgment}

This work was partially supported by the European Commission through the Horizon 2020 research and innovation program under grants 769355 (PIXEL).

\bibliographystyle{IEEEtran}
\bibliography{IEEEabrv,IEEEexample}

\thanks{
\textcopyright 2020 IEEE. Personal use of this material is permitted. Permission from IEEE must be obtained for all other uses, in any current or future media, including reprinting/republishing this material for advertising or promotional purposes, creating new collective works, for resale or redistribution to servers or lists, or reuse of any copyrighted component of this work in other works.
}

\end{document}